\documentclass[10pt, a4paper]{article}
\usepackage{float} 
\usepackage{multirow}
\usepackage[final]{lrec2026} 

\usepackage{graphicx}
\usepackage{subcaption}

\title{Injecting Structured Biomedical Knowledge into Language Models: Continual Pretraining vs. GraphRAG}

\name{
Jaafer Klila$^{\ast\dagger}$, Sondes Bannour Souihi$^{\ast}$, Rahma Boujelben$^{\dagger}$ \\
{\bf \large Nasredine Semmar$^{\ast}$, Lamia Hadrich Belguith$^{\dagger}$}
}

\address{
$^{\ast}$Université Paris-Saclay, CEA, List \\ Palaiseau, France \\
klilajaafer@gmail.com \\
\{sondes.souihi, nasredine.semmar\}@cea.fr \\
$^{\dagger}$University of Sfax \\ Sfax, Tunisia \\
\{rahma.boujelbane, lamia.belguith\}@fsegs.usf.tn
}

\abstract{
The injection of domain-specific knowledge is crucial for adapting language models (LMs) to specialized fields such as biomedicine. While most current approaches rely on unstructured text corpora, this study explores two complementary strategies for leveraging structured knowledge from the UMLS Metathesaurus: (i) Continual pretraining that embeds knowledge into model parameters, and (ii) Graph Retrieval-Augmented Generation (GraphRAG) that consults a knowledge graph at inference time. We first construct a large-scale biomedical knowledge graph from UMLS (3.4 million concepts and 34.2 million relations), stored in Neo4j for efficient querying. We then derive a ~100-million-token textual corpus from this graph to continually pretrain two models: BERTUMLS (from BERT) and BioBERTUMLS (from BioBERT). We evaluate these models on six BLURB (Biomedical Language Understanding and Reasoning Benchmark) datasets spanning five task types and evaluate GraphRAG on the two QA (Question Answering) datasets (PubMedQA, BioASQ). On BLURB tasks, BERTUMLS improves over BERT, with the largest gains on knowledge-intensive QA. Effects on BioBERT are more nuanced, suggesting diminishing returns when the base model already encodes substantial biomedical text knowledge. Finally, augmenting LLaMA 3-8B with our GraphRAG pipeline yields over than 3 points accuracy on PubMedQA and 5 points on BioASQ without any retraining, delivering transparent, multi-hop, and easily updated knowledge access. We release the processed UMLS Neo4j graph to support reproducibility.
 \\ \newline 
}

\begin{document}
\pagestyle{empty}
\maketitleabstract

\section{Introduction}

Language Models (LMs) have rapidly transformed the field of Natural Language Processing (NLP), enabling machines to understand and generate human language with remarkable fluency. In the biomedical field, early Pretrained Language Models (PLMs), such as  BioBERT \cite{lee2020biobert}, ClinicalBERT \cite{alsentzer2019publicly} and PubMedBERT \cite{gu2021domain} extended BERT \cite{devlin2019bert} by pretraining on large-scale biomedical corpora, including PubMed \footnote{\url{https://pubmed.ncbi.nlm.nih.gov/download/}}, PMC \footnote{\url{https://pmc.ncbi.nlm.nih.gov/}}, and MIMIC-III \cite{johnson2016mimic}, yielding strong gains through richer contextualization of biomedical terminology and discourse.

More recently, Large Language models (LLMs) such as GPT-4 \cite{achiam2023gpt}, DeepSeek \cite{liu2024deepseek}, and Llama 4 models \cite{meta2025llama4card}, have reported competitive or even human-level performance on a range of general benchmarks, further raising expectations for domain adaptation.

Despite these advances, several challenges persist in knowledge-intensive domains. First, because LMs \footnote{We use the term "LMs" to collectively refer to both PLMs and LLMs.} are trained on static snapshots of data, they often lack up-to-date factual knowledge and struggle to reflect recent findings \cite{he2022rethinking,melnyk2021grapher}. Second, they can generate confident but incorrect outputs \cite{ji2023survey,bang2023multitask}, with hallucinations remaining a prominent failure mode \cite{huang2025survey}. Third, model predictions are typically opaque: it is difficult to trace, verify, or cite the sources underpinning an answer \cite{pan2024unifying}. Finally, because LMs are fundamentally probabilistic next-token predictors, their access to structured or semantic relations, such as synonymy, hierarchical typing, and multi-hop associations among biomedical entities, remains indirect when relying solely on unstructured text. This gap is particularly limiting for tasks like biomedical question answering, where accurate reasoning often depends on chaining relations across curated resources.

These limitations argue for methods that move beyond continued pretraining on raw text and instead inject structured, verifiable knowledge directly into the modeling pipeline. In this paper, we investigate two complementary strategies grounded in the UMLS Metathesaurus \footnote{\url{https://uts.nlm.nih.gov/uts/umls/home}}. The first is a parametric strategy, continual pretraining on a textualized graph-to-text corpus, so that knowledge becomes embedded in the model’s parameters and is always available at inference. The second is a non-parametric strategy, Graph Retrieval-Augmented Generation (GraphRAG) \cite{han2024retrieval}, that keeps knowledge external in a Neo4j \footnote{\url{https://neo4j.com/}} graph and retrieves the relevant subgraph at inference, affording explainability (inspectable paths), multi-hop reasoning through traversal, and rapid updates by refreshing the graph rather than retraining the model.

Concretely, we construct a large-scale UMLS-derived biomedical knowledge graph  publicly available at: \url{https://github.com/jaaferklila/UMLS_knowledge_graph}
and derive a ~100-million-token textualized corpus from it. We then (i) perform continual pretraining to obtain BERTUMLS (from BERT) and BioBERTUMLS (from BioBERT), and (ii) implement a GraphRAG pipeline with LLaMA 3-8B to consult the graph at inference. Evaluation covers six BLURB datasets \cite{gu2021domain}  spanning five task types, with GraphRAG assessed on the two knowledge-intensive QA datasets (PubMedQA, BioASQ). The results preview the distinct benefits of each strategy: continual pretraining yields the largest gains for a general-domain base, whereas effects on the domain-specialized base are more nuanced, suggesting diminishing returns when biomedical knowledge is already well encoded. In contrast, GraphRAG improves performance across both QA datasets without retraining, promoting transparent, multi-hop, and easily updated knowledge access.
The remainder of the paper is organized as follows. In Section 2, we survey previous works addressing the task of biomedical knowledge injection. In Section 3, we detail UMLS graph construction and graph-to-text textualization, and we present the continual-pretraining setup and the GraphRAG pipeline. The experimental results are reported and discussed in Section 4. Finally, we present in Section 5 the conclusion and future work.
\section{Related Work}
Injecting domain-specific knowledge is widely recognized as a key lever for improving the reliability and usefulness of language models in specialized settings. Recent surveys map four main adaptation families: continual pre-training on domain-specific corpora, knowledge injection through modular Adapters, Retrieval-Augmented Generation (RAG) for dynamic information access, and Prompt Optimization methods \cite{song2025injecting}.
In the biomedical field, continual pretraining remains a strong baseline: PMC-LLaMA adapts a general LLaMA on a large corpus of papers and textbooks \cite{wu2024pmc}, while BioMedLM trains 2.7B parameters exclusively on PubMed \cite{bolton2024biomedlm}. These text-centric approaches improve fluency and coverage, but they typically encode relations like synonymy, typing and causal or therapeutic links only implicitly and require costly retraining to stay current.
A complementary thread focuses on making structure explicit. OntoTune \cite{liu2025ontotune} proposes ontology-driven self-training to align LLMs with hierarchical resources such as SNOMED CT \cite{schulz2008snomed}, using in-context learning to detect and repair conceptual gaps.
In parallel, BiomedRAG \cite{li2025biomedrag} demonstrates the value of non-parametric access to knowledge by retrieving chunked biomedical documents at inference, yielding gains across several biomedical NLP tasks. Yet, even here, most systems still deliver knowledge through text, which can dilute entity-level relations, complicate multi-hop reasoning, and hinder explanation. Moreover, updating knowledge often remains tied to retraining pipelines or large-scale re-indexing.

This work addresses those gaps by operationalizing structured knowledge as a graph and by comparing two complementary injection strategies on the same underlying biomedical resource. We build a UMLS-derived knowledge graph in Neo4j and, from that single source, pursue (i) a parametric path that performs continual pretraining on a textualized graph-to-text corpus, tested on both a general model (BERT) and a specialized one (BioBERT), and (ii) a non-parametric path that implements GraphRAG, retrieving subgraphs for a generative LLaMA-family model at inference. This design isolates the role of access mode, embedded vs. retrieved, while directly targeting explainability (inspectable evidence chains), multi-hop reasoning (graph traversal), and freshness (graph refreshes rather than model retraining). In contrast to prior work that either relies purely on biomedical text or introduces ontology signals without a side-by-side comparison to graph-centric retrieval, our study offers a controlled view of how structured knowledge should be injected to maximize downstream utility.

\section{Knowledge Injection Methodology}

\begin{figure*}[htbp]
    \centering
    \includegraphics[scale=0.4, keepaspectratio]{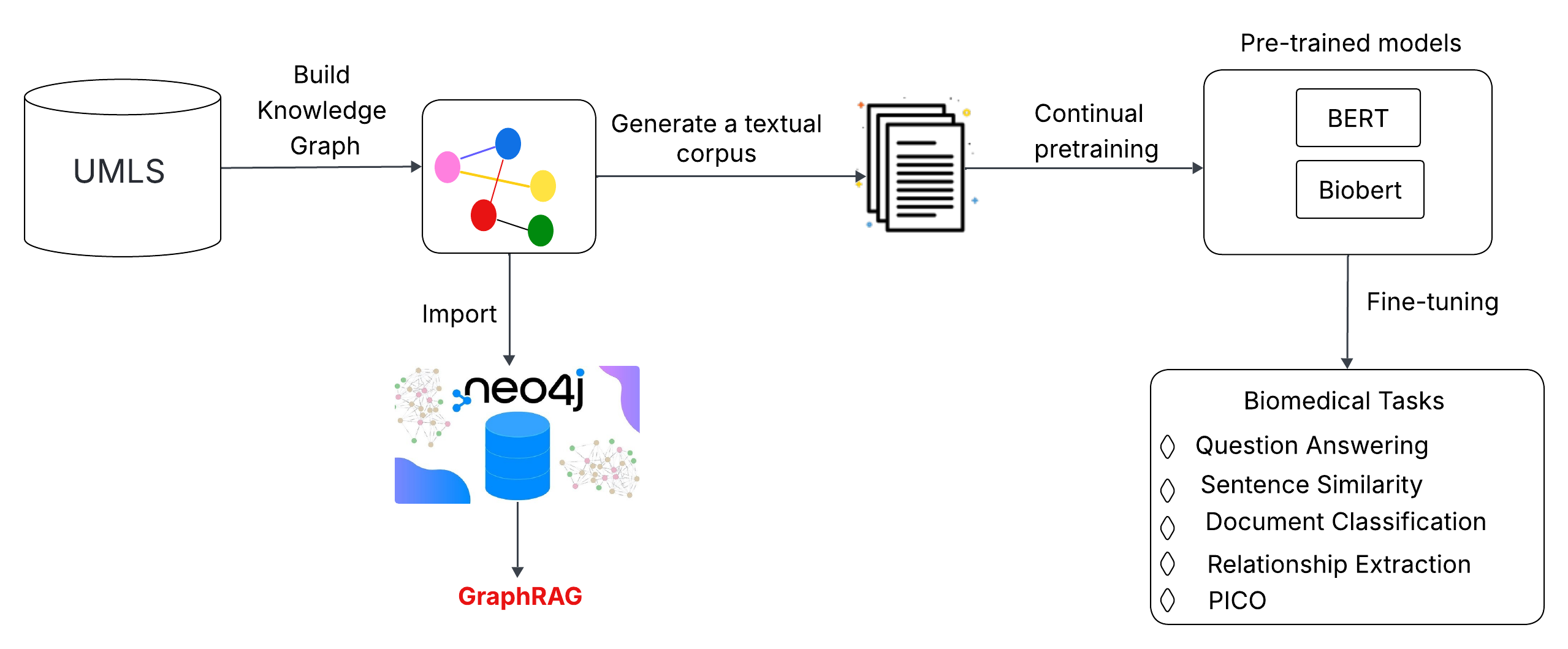} 
    \caption{Overview of the UMLS-Based Knowledge Injection Pipeline: Continual Pretraining on BERT/BioBERT and GraphRAG.}
    \label{fig:training_architecture}
\end{figure*}

We investigate two complementary avenues for injecting structured biomedical knowledge into language models. First, we introduce BioBERTUMLS and BERTUMLS by continuing pretraining with knowledge derived from the UMLS knowledge graph followed by fine-tuning on downstream biomedical NLP tasks to evaluate their effectiveness in domain-specific applications. Second, we evaluate the value of structured evidence at inference via a GraphRAG framework built on the LlaMA family for question answering. The overall architecture of our pipeline is shown in Figure~\ref{fig:training_architecture}.

Our knowledge source is the UMLS-2024AA Metathesaurus, the largest component of UMLS, which consolidates biomedical concepts, their names, semantic types, and inter-concept relations. We construct a knowledge graph from this release and use it both to generate the textualized corpus for continual pretraining and to power graph-based retrieval at inference time. This single-source design ensures that comparisons between parametric (pretraining) and non-parametric (GraphRAG) strategies reflect differences in access mode rather than differences in content. 

Concretely, Section 3.1 details the UMLS background. Section 3.2 describes graph construction. Section 3.3 presents knowledge injection via continual pretraining and Section 3.4 introduces our GraphRAG setup with LLaMA over the Neo4j-hosted graph for biomedical QA.

\subsection{Unified Medical Language System (UMLS)} 
The Unified Medical Language System (UMLS)~\cite{UMLS2024} is a comprehensive aggregation of biomedical vocabularies and ontologies maintained by the U.S. National Library of Medicine. The UMLS-2024AA Metathesaurus integrates over 200 source vocabularies, offering a unified layer for mapping concepts across heterogeneous terminologies.
UMLS is organized around three complementary resources:

\begin{itemize}
    \item The \textbf{Metathesaurus}, which forms the core component, contains information about biomedical concepts, their various names (terms) from the source vocabularies, and the relationships between them.
    \item The \textbf{Semantic Network} provides a consistent categorization of all concepts in the Metathesaurus into a set of broad \textit{Semantic Types} (e.g., `Disease or Syndrome', `Pharmacologic Substance') and defines the permissible relationships between these types.
    \item The \textbf{SPECIALIST Lexicon} and associated lexical tools provide syntactic information for English biomedical terms, supporting natural language processing tasks such as part-of-speech tagging and morphological analysis.
\end{itemize} 

The Metathesaurus is distributed as relational-style RRF tables, in this work we rely primarily on \texttt{MRCONSO.RRF} (concept names, sources, identifiers), \texttt{MRDEF.RRF} (definitions), \texttt{MRSTY.RRF} (semantic types), and \texttt{MRREL.RRF} (relationships between concepts).

\subsection{Knowledge Graph Construction}
\label{ssec:kg_construction}

A Knowledge Graph (KG) is a semantic representation of real-world entities and the relationships among them, typically modeled as triples (h,r,t), where h is the head entity, t the tail entity, and r the relation. Formally, a KG can be written as:
\[
KG = \{(h, r, t)\}
\]

\begin{figure}[!ht]
\begin{center}
\includegraphics[width=\columnwidth]{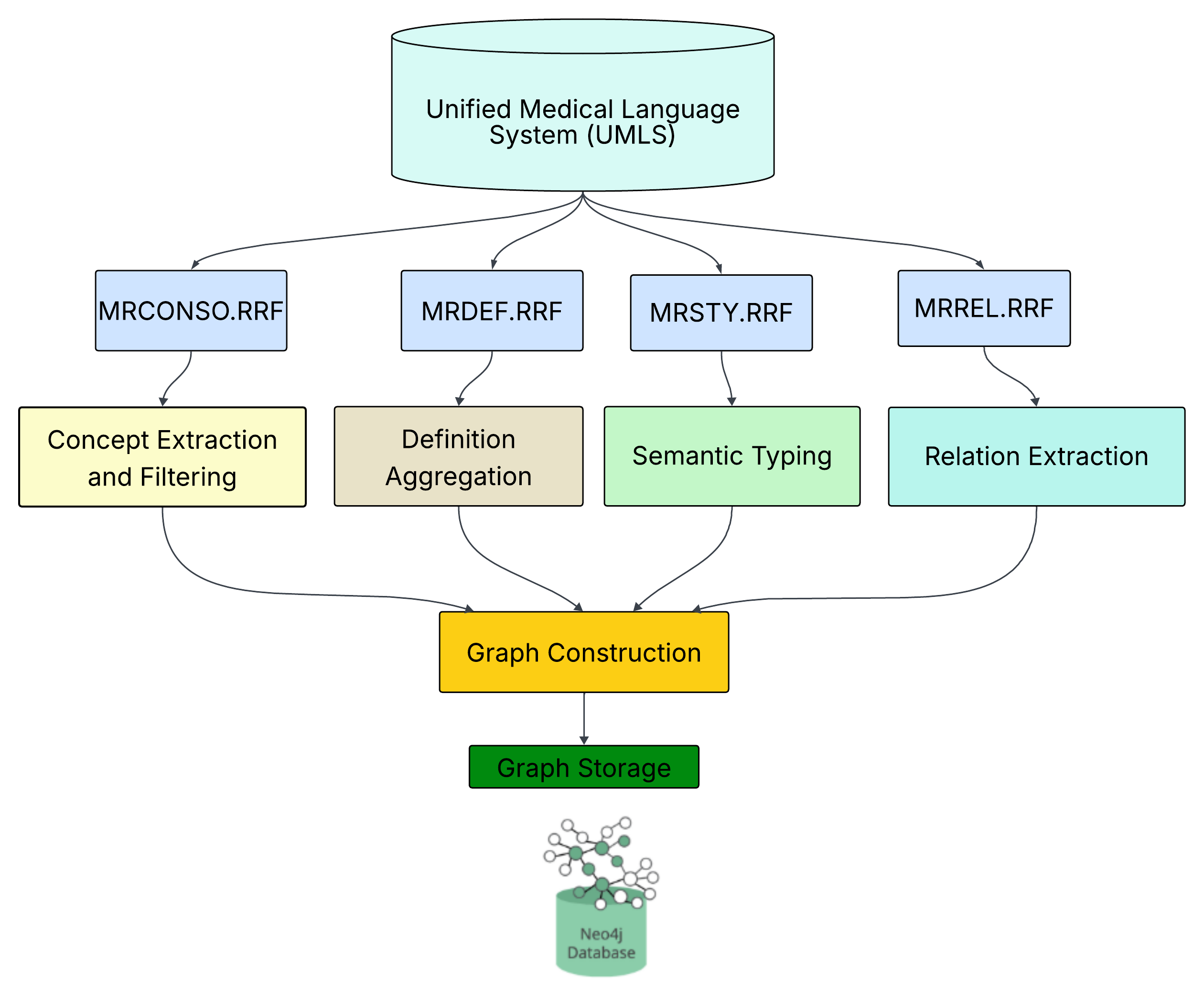}
\caption{Pipeline for constructing a biomedical
knowledge graph from UMLS Metathesaurus.}
\label{fig.3}
\end{center}
\end{figure}

The pipeline used to construct our KG is shown in Figure~\ref{fig.3}. We begin from the core UMLS tables (\texttt{MRCONSO}, \texttt{MRDEF}, \texttt{MRSTY}, and \texttt{MRREL)} and proceed as follows.

\noindent\textbf{Concept Extraction and Filtering.} From the \texttt{MRCONSO} table, which contains approximately 17 million entries, we extracted concept names. Because concepts may have multiple names across languages and vocabularies, we filtered to retain only English (LAT='ENG') entries from English-language sources to ensure linguistic consistency.

\noindent\textbf{Definition Aggregation.} The \texttt{MRDEF} table provided 466,842 concept definitions from sources such as MeSH and CSP. After processing, we identified 287,972 unique English concepts associated with 339,341 definitions; 32,219 concepts had more than one definition. For example, concept \texttt{C0018798} (Congenital Heart Defects) was associated with 24 distinct definitions. To preserve this richness, we concatenated all definitions for a given concept into a single field with a separator for downstream access.

\noindent\textbf{Semantic Typing.} Semantic types for each concept were sourced from the \texttt{MRSTY} table, which links Concept Unique Identifiers (CUIs) to one or more of the 127 distinct semantic types (e.g., `Disease or Syndrome', `Gene or Genome'). This provides a high-level categorization for all entities in our KG.

\noindent\textbf{Relation Extraction.} The relational structure derives from \texttt{MRREL}, which contains 62.9 million relations spanning 1,035 unique relationship types (RELA values). After filtering to English concepts and removing self-relations (where CUI1 = CUI2), the final set comprised 34.2 million relationships between 3.39 million concepts, covering 1,005 distinct types. When a RELA value was missing, we substituted the more general REL value to avoid data loss; this heuristic recovered over 9 meaningful relationships for the final graph.

\noindent\textbf{Graph Construction.} After data extraction from UMLS Metathesaurus we performed joins between four tables to associate each concept with its names, synonyms, relationships, and semantic types.
The resulting data was then stored as a graph in a Neo4j database.

\noindent\textbf{Graph Storage.} The resulting Knowledge Graph, comprising 3,389,266 concepts and 1,005 unique relationship types, was stored in a Neo4j graph database. This storage solution was selected to facilitate efficient querying and, crucially, for straightforward integration with GraphRAG applications. 
\subsection{Knowledge Injection via Continual Pretraining }
Continual pretraining updates an already pretrained language model by resuming its original objective on a new knowledge-rich corpus, with the goal of injecting domain knowledge directly into the model parameters rather than training from scratch. In our setting, the corpus is derived from a structured biomedical resource (UMLS). Our evaluation is designed to answer three questions: (i) does a biomedical model such as BioBERT still benefit from this knowledge-injection procedure, (ii) does a general model such as BERT obtain a measurable gain, and (iii) does structured knowledge injection yield greater gains than training on specialized biomedical corpora alone (i.e., BioBERT-style domain pretraining)?.

\subsubsection{ Textual Corpus Construction}
Inspired by the approach described in \cite{piat2023integration}, we convert the Knowledge Graph into text so that structured facts can be consumed by a language model while preserving their original form. We refer to this step as textualization. The textualization uses two ingredients from UMLS:
\begin{itemize}
    \item \textbf{Factual triples} (h,r,t) where h and t are concepts (CUIs) and r is the labeled edge connecting them;
    \item \textbf{Concept definitions} obtained from \texttt{MRDEF} (English). We include 285,818 unique English definitions.
\end{itemize}

\begin{figure}[!ht]
\begin{center}
\includegraphics[width=\columnwidth]{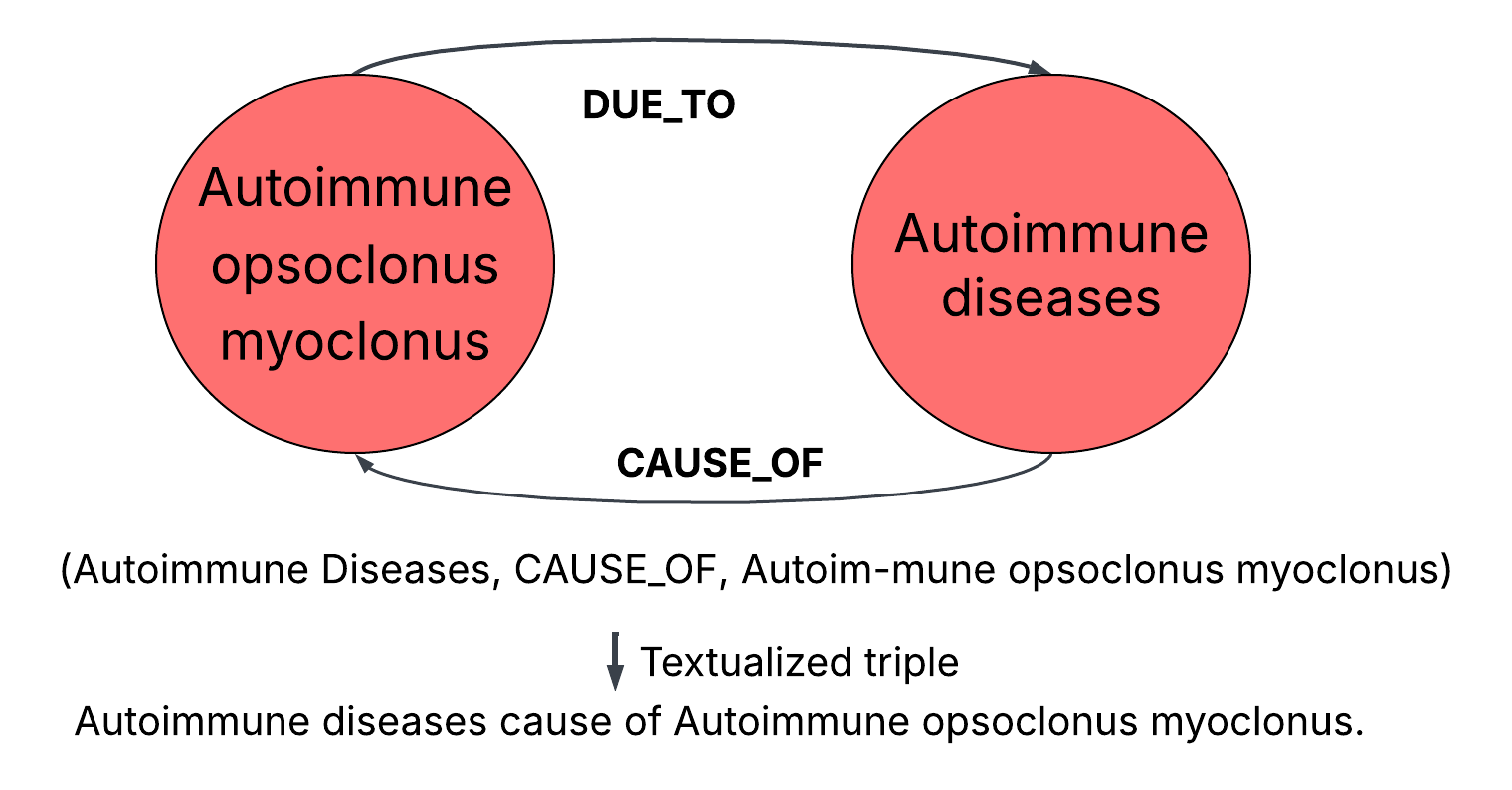}
\caption{Example of generating a textualized triple from the KG.}
\label{fig.22}
\end{center}
\end{figure}

For each triple (h,r,t), we produce a compact fragment that states the relation label r between the two concepts. We keep concept surface forms and relation labels consistent with UMLS. A running example is shown in Figure~\ref{fig.22}.

For example, for the triple (Autoimmune opsoclonus myoclonus , CAUSE\_OF, Autoimmune diseases), the system produces the fragment:

\begin{quote}
``Autoimmune diseases cause of Autoimmune opsoclonus myoclonus.''
\end{quote}

These natural-language textualized triples are designed to be unambiguous and easily tokenizable while mirroring the underlying graph structure.

For each concept node, we build a short, self-contained block that aggregates:
\begin{itemize}
    \item the concept's name and synonyms (from \texttt{MRCONSO});
    \item its semantic type(s) (from \texttt{MRSTY});
    \item the set of outgoing/incoming relation fragments (from \texttt{MRREL}), each in the format above.
\end{itemize}

This structure gives the model repeated, consistent exposure to how a concept is situated in the graph (neighbors and relation labels) without relying on prose. Concept definitions (\texttt{MRDEF}) provide curated, human-readable descriptions that complement relation fragments. When a concept has multiple definitions, we concatenate them so that all available descriptions remain accessible in downstream use.

The resulting corpus comprises approximately 100 million words, combining concept-centric blocks (names, semantic types, definitions) and the full set of relation fragments. This forms a knowledge-grounded resource tailored to continual pretraining. The expectation is that this will strengthen the model’s internal representation of biomedical entities and relations, which we subsequently test on downstream tasks.

\subsection{Knowledge Injection via Graph Retrieval-Augmented Generation }
\label{ssec:graph_rag}

Graph Retrieval-Augmented Generation (GraphRAG) augments large language models with an external knowledge graph to produce more complete and better-grounded answers~\cite{edge2024local}. In this setup, the model’s parametric knowledge is complemented with explicit, verifiable structure drawn from the graph, which can be inspected, updated, and constrained independently of the model.

\begin{figure}[!ht]
\begin{center}
\includegraphics[width=\columnwidth]{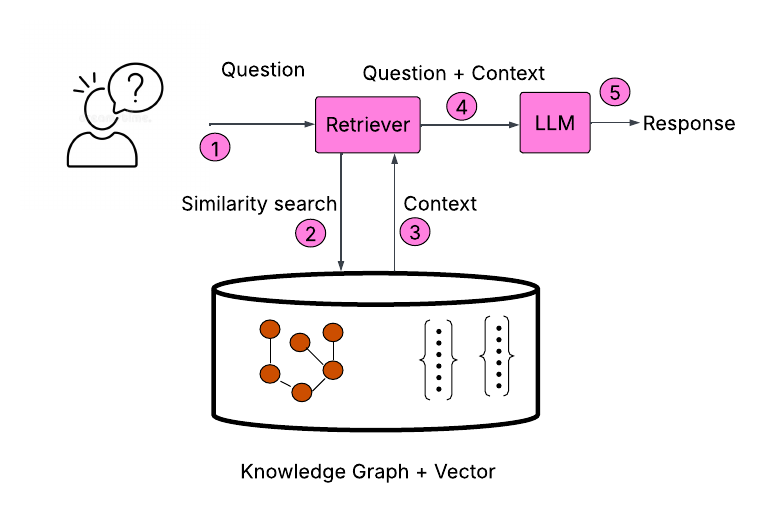}
\caption{GraphRAG pipeline.}
\label{fig.222}
\end{center}
\end{figure}

As shown in Figure \ref{fig.222}, a user question is first submitted to a vector similarity search engine to retrieve relevant text passages. Subsequently, these passages seed a graph traversal step that extracts a subgraph of related entities and connections. This enriched context is then provided to an LLM to generate a  response based on both the context and the LLM’s prior knowledge.

We adapt GraphRAG to the biomedical domain by using the Unified Medical Language System (UMLS) knowledge graph, stored and queried using a Neo4j database, as our primary, verifiable source of structured knowledge. For the initial retrieval phase, we employ model2vec \cite{minishlab2024word2vec} to identify semantically relevant text passages from a biomedical graph. The retrieved structured subgraph, transformed into textual triples (h, r, t), is then integrated into the prompt as contextual information for the Llama-3.1-8B \cite{grattafiori2024llama} model to generate informed responses. Finally, we specifically design and evaluate this pipeline to assess its efficacy on biomedical question-answering tasks.
\section{Experimentation}
Our experiments compare two complementary strategies for injecting domain knowledge derived from UMLS-2024AA:
(1) parametric injection via continual pretraining (yielding BERTUMLS and BioBERTUMLS), and
(2) non-parametric injection via GraphRAG with LLaMA 3-8B over a Neo4j instance of the UMLS graph.
Both strategies rely on the same UMLS snapshot to avoid confounding content differences: the snapshot is used to build the graph consulted at inference (GraphRAG) and to produce the textual corpus used for continual pretraining. We evaluate on six biomedical datasets spanning five task types and report each dataset’s official metric. To reflect variability, we run five independent runs and report mean standard deviation.
\subsection{Tasks \& Datasets}
We evaluate on BLURB\footnote{\url{https://microsoft.github.io/BLURB/}} benchmark, covering five biomedical task families with six datasets. We use each dataset’s official splits and official evaluation metric (Table~\ref{tab:biomedical_datasets}).

\begin{table*}[htbp]
\resizebox{\textwidth}{!}{
\centering

\begin{tabular}{l l r r r l r}
\hline
Dataset & Task & Train & Dev & Test & Evaluation Metrics & Number of Classes \\
\hline
EBM PICO & PICO & 339,167 & 85,321 & 16,364 & Macro F1 word-level & 3 \\
BIOSSES & Sentence Similarity & 64 & 16 & 20 & Pearson & N/A (labels continus) \\
HoC & Document Classification & 1,295 & 186 & 371 & Average Micro F1 & 10\\
DDI & Relation Extraction & 25,296 & 2,496 & 5,716 & Micro F1 & 18\\
PubMedQA & Question Answering & 450 & 50 & 500 & Accuracy & 3 \\
BioASQ & Question Answering & 670 & 75 & 140 & Accuracy & 2 \\
\hline
\end{tabular}
}
\caption{Summary of the biomedical datasets used with the number of classes and evaluation metrics.}
\label{tab:biomedical_datasets}
\end{table*}
\textbf{ Relation Extraction.} Relation Extraction (RE) is a critical NLP task that involves identifying and classifying relationships between entities within text \cite{zhong2020frustratingly, lai2023biorex}. We evaluate RE on the GAD dataset~\cite{becker2004genetic}.

\textbf{ Question Answering.} Automatic question answering (QA) systems involve developing technologies that automatically provide answers to questions asked by humans in a specific domain. These systems have been successfully deployed in various domains such as search engines and chatbots \cite{jin2022biomedical}. We evaluate QA on 2 Biomedical QA datasts, PubMedQA~\cite{jin2019pubmedqa}  and BioASQ~\cite{baker2016automatic}.

\textbf{Sentence Similarity.} Sentence Similarity is the task of determining how similar two documents, sentences or texts are based on their meaning. Sentence similarity models transform input texts into semantic vector representations (embeddings), which can then be used to quantify the proximity between them. We evaluate sentence similarity on the BIOSSES dataset~\cite{souganciouglu2017biosses}.

\textbf{Document Classification.} We utilize the Hallmarks of Cancer (HoC) corpus~\cite{hanahan2000hallmarks} for evaluation, a resource designed for multi-label sentence classification. It consists of 1,852 PubMed abstracts where each sentence has been manually annotated by experts with any number of labels from a hierarchical taxonomy of 37 cancer hallmarks.

\textbf{PICO.} The EBM PICO corpus~\cite{nye2018corpus} contains approximately 5,000 medical abstracts describing clinical trials. Each abstract has been meticulously annotated to identify PICO elements: Population (e.g., 'diabetics'), Intervention ('insulin'), Comparator ('placebo'), and Outcome ('blood glucose levels').
\subsection{Baselines}
We use three baselines in our experiments:
\begin{itemize}
    \item BERT is the general-domain baseline with no biomedical specialization and no access to UMLS; it is evaluated on all six datasets
    \item BioBERT is the biomedical baseline pretrained on large specialized corpora (no UMLS structure); it is likewise evaluated on all six datasets. BioBERT serves as the specialized-corpus control in our study. It lets us establish how far domain text alone can push performance across BLURB tasks and test whether further knowledge injection still helps a model that already encodes extensive biomedical regularities.
    \item LLaMA 3-8B (without external knowledge) is the generative baseline used only for the QA tasks (PubMedQA, BioASQ); it provides the point of comparison for its GraphRAG variant to isolate the effect of inference-time UMLS access.
\end{itemize}
\subsection{Continual Pretraining}
\subsubsection{Setup}
For continual pre-training, we initialize both models (BERT and BioBERT) with their existing pre-trained weights. We inject UMLS knowledge parametrically by resuming the original pretraining objective (Masked Language Modeling (MLM)) on the UMLS-derived textual corpus. We use the following hyperparameters: a batch size of 512, a maximum sequence length of 512, and the AdamW optimizer with a learning rate of 2e-5. The models are trained for 1 epoch on our corpus.
\subsubsection{Results and Interpretation}
The results of knowledge injection via continual pretraining for the general domain model are shown in Table~\ref{tab:bert_results} and for the specialized domain model in Table~\ref{tab:biobert_results}.
\begin{table}[H]
\centering
\begin{tabular}{|l|c|c|}
\hline
 Dataset & BERT & BERTUMLS  \\
\hline
PubMedQA &  $50.72 \pm 0.17  $&  $\textbf{53.00} \pm 0.0 $ \\
\hline
BioASQ   &$ 66.57 \pm 0.93  $&  $\textbf{71.57} \pm 1.17 $ \\
\hline
EBM PICO &$ 71.95\pm 0.06 $ & $\textbf{72.18}\pm 0.05 $\\
\hline
 DDI &$ 77.99 \pm 0.19 $&  $\textbf{78.67}\pm 0.27  $\\
\hline
BIOSSES & $82.31\pm 4.74 $& $\textbf{82.54}\pm 1.14 $ \\
\hline
 HoC & $\textbf{80.60}\pm 0.35 $ &$ 80.57 \pm 0.88 $\\
\hline
\end{tabular}
\caption{BERT vs BERTUMLS (ours) test results (\%) across biomedical NLP tasks. Evaluation metrics are listed in Table \ref{tab:biomedical_datasets}.}
\label{tab:bert_results}
\end{table}
\begin{table}[h]
\centering
\begin{tabular}{|l|c|c|}
\hline
Dataset & BioBERT & BioBERTUMLS \\
\hline
PubMedQA & $52.80 \pm 0.0$ & $\textbf{54.20} \pm 0.0$ \\
\hline
BioASQ & $\textbf{79.71} \pm 0.38$ & $75.57 \pm 1.25$ \\
\hline
EBM PICO & $73.56 \pm 0.01$ & $\textbf{73.66} \pm 0.04$ \\
\hline
DDI & $\textbf{81.64} \pm 0.19$ & $80.97 \pm 0.33$ \\
\hline
BIOSSES & $88.37 \pm 2.02$ & $\textbf{90.36} \pm 0.45$ \\
\hline
HoC & $82.82 \pm 0.22$ & $\textbf{83.03} \pm 0.25$ \\
\hline
\end{tabular}
\caption{BioBERT vs BioBERTUMLS (ours) test results (\%) across biomedical NLP tasks. Evaluation metrics are listed in Table \ref{tab:biomedical_datasets}.}
\label{tab:biobert_results}
\end{table}
We observe that BERTUMLS, continually pretraining BERT on a UMLS-derived corpus, outperforms BERT which was only pretrained on the general domain across five out of six biomedical NLP tasks, with only the document classification task (HoC dataset) showing a negligible decrease of 0.04\%, which is statistically insignificant. The injection of UMLS knowledge through continual pretraining  demonstrates its effectiveness for question answering tasks, with BioASQ and PubMedQA exhibiting the most substantial improvements of approximately 7.6\% and 4.5\%, respectively. Moderate gains are observed in relation extraction (DDI), EBM PICO, and sentence similarity (BIOSSES), all under 1\%, indicating that the benefit of domain knowledge is less pronounced in tasks that rely more on syntactic or surface-level features. Overall, these findings highlight that continual pretraining with UMLS knowledge enhances the model’s ability to leverage biomedical concepts, providing meaningful performance gains, especially in tasks that demand deep domain understanding.
On the other hand, injecting UMLS knowledge into the specialized BioBERT model does not yield consistent performance gains across all tasks. For instance, BioBERTUMLS improves on PubMedQA by 2.65\% and on BIOSSES by 2.26\%. It shows smaller gains on EBM PICO of 0.14\% and on HoC of 0.25\%. In contrast, performance declines on BioASQ by 5.19\% and on DDI by 0.82\%. This behavior can be attributed to the fact that BioBERT is already pretrained on an extensive corpus of biomedical text (approximately 3.4B words), whereas our UMLS-based corpus is significantly smaller (about 100M words, roughly 3\% of BERT and 2.2\% of BioBERT pretraining corpus) (see Table~\ref{tab:pretraining_corpus}). This limits the marginal benefit of additional domain-specific knowledge.  
\begin{table}[h]
\centering
\begin{tabular}{|l|c|c|}
\hline
Model & Number of words & Domain \\
\hline
BERT & 3.4B & General \\
BioBERT & 4.5B & Biomedical \\
BERTUMLS & 100M & Biomedical \\
BioBERTUMLS & 100M & Biomedical \\
\hline
\end{tabular}
\caption{Pretraining corpora of biomedical and general language models.}
\label{tab:pretraining_corpus}
\end{table}
Overall, these results suggest that continual pretraining with UMLS knowledge is particularly beneficial for general-domain models, while its effect on specialized models is more nuanced and task-dependent.
BERTUMLS delivers a compelling return on UMLS-based knowledge injection: despite being continued on a comparatively small (100M-word) corpus, it closes much of the gap to BioBERT and even surpasses it on PubMedQA. On BioASQ and the non-QA tasks, BioBERT still leads, but the margins are substantially narrower than one would expect given BioBERT’s orders-of-magnitude larger pretraining footprint. In other words, targeted, structured knowledge can lift a general model into the neighborhood of a domain-specialized one, and, for certain QA regimes, nudge it ahead, without billions of extra tokens or costly retraining. This positions BERTUMLS as a strong, compute-efficient baseline for knowledge-intensive biomedical tasks and a promising foundation for hybrid pipelines (e.g., pairing parametric UMLS injection with retrieval-time graph context) to further erode any remaining gaps.
\subsection{GraphRAG}
\subsubsection{Setup}
We inject UMLS knowledge non-parametrically at inference via GraphRAG. A question triggers vector similarity search (model2vec) to retrieve the relevant subgraph from the UMLS graph stored in Neo4j. The subgraph is textualized as triples (h,r,t) and inserted as context for LLaMA 3-8B to generate an answer. No retraining is performed. Evaluation is on PubMedQA and BioASQ with accuracy. 
This setup isolates the effect of inference-time access to the same UMLS knowledge: transparent evidence (inspectable subgraphs), multi-hop reasoning through explicit relations, and easy knowledge updates via graph refreshes.
\subsubsection{Results and Interpretation}
\begin{figure}[htbp]
    \centering
    \includegraphics[width=\linewidth, keepaspectratio]{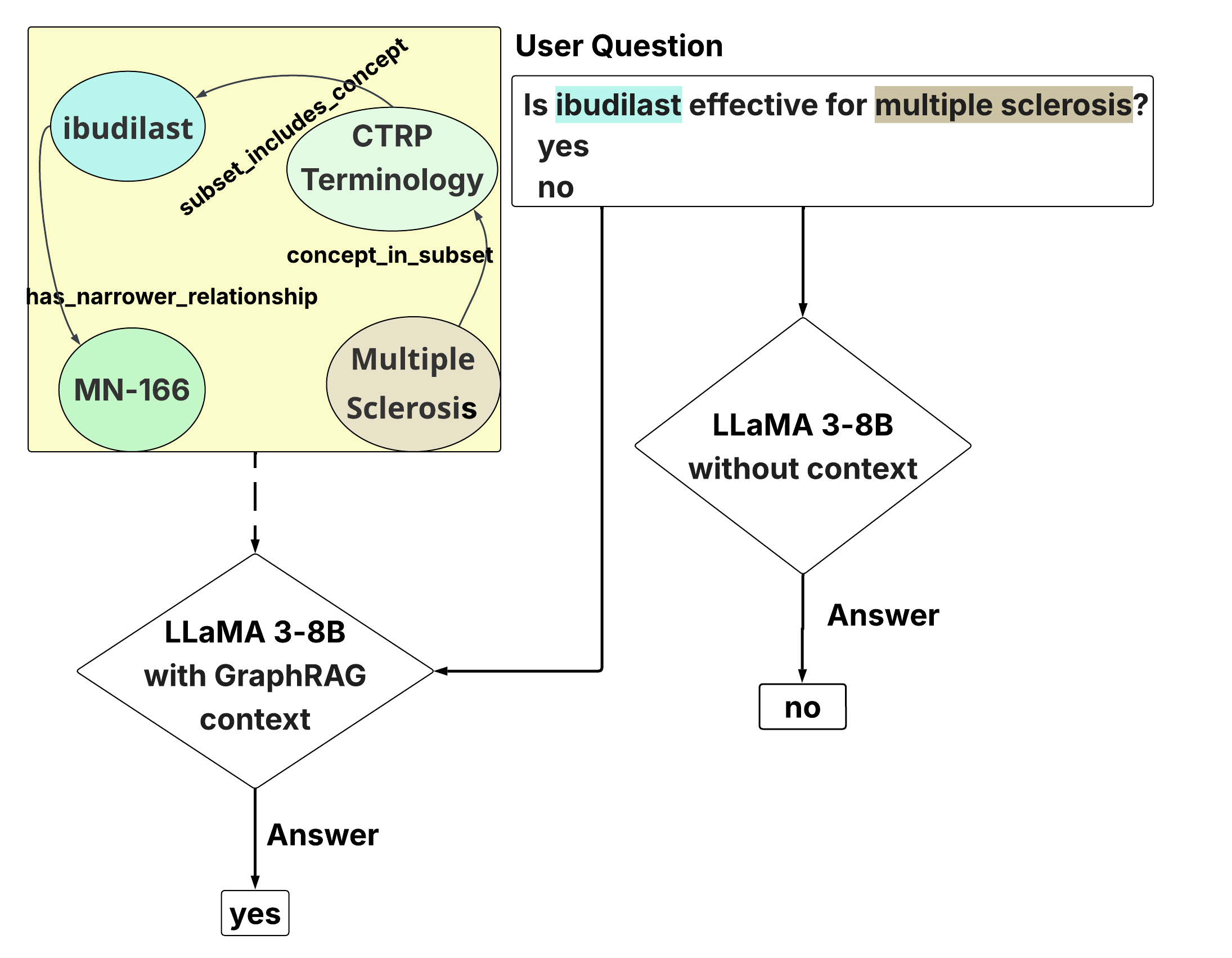}
    \caption{A real-world biomedical question from BioASQ and the corresponding subgraph extracted from our Neo4j knowledge graph. Using a LLama model alone often yields an incorrect answer (‘No’). By injecting structured biomedical knowledge from the UMLS graph as context, the model correctly answers ‘Yes’.}
    \label{fig:graph1211}
\end{figure}

Table \ref{tab:llama_rag_results} shows that augmenting the LLaMA 3-8B model with our GraphRAG pipeline consistently improves performance across both datasets with no retraining. Specifically, we observe a 6.25\% increase in accuracy on PubMedQA and a 6.67\% increase on BioASQ. These improvements highlight the effectiveness of integrating structured, retrievable knowledge from biomedical graphs, enabling the model to answer questions more accurately. 
LLaMA with GraphRAG even outperforms both biomedical pretraining models, BERTUMLS and BioBERTUMLS.
Relative to parametric injection, GraphRAG remains competitive or better on QA when measured in relative terms. On PubMedQA, GraphRAG improves over BERTUMLS by 2.64\% and over BioBERTUMLS by 0.37\%. On BioASQ, GraphRAG improves over BERTUMLS by 11.78\%, over BioBERTUMLS by 5.86\%, and even edges the BioBERT baseline by 0.36\%. These gains isolate the contribution of inference-time access to structured knowledge, since the underlying LLM is unchanged and only the prompt is augmented with a focused UMLS subgraph.
\begin{table}[h]
\centering
\small
\begin{tabularx}{0.9\columnwidth}{|l|X|X|}
\hline
Dataset & \centering LLaMA 3-8B & \centering LLaMA 3-8B + GraphRAG \tabularnewline
\hline
PubMedQA & \centering 51.2 & \centering \textbf{54.4} \tabularnewline
\hline
BioASQ & \centering 75.0 & \centering \textbf{80.0} \tabularnewline
\hline
\end{tabularx}
\caption{Comparison of accuracy (\%) on PubMedQA and BioASQ between LLaMA 3-8B and our GraphRAG-enhanced model.}
\label{tab:llama_rag_results}
\end{table}
Figure~\ref{fig:graph1211} illustrates a real-world biomedical question from BioASQ along with the corresponding subgraph extracted from our Neo4j knowledge graph. The user asks: "Is ibudilast effective for multiple sclerosis?" Without external knowledge, LLaMA 3-8B answers no. With GraphRAG, the prompt includes a compact UMLS subgraph that links "ibudilast" to "CTRP Terminology" and, through subset\_includes\_concept / concept\_in\_subset relations, connects that terminology node to "multiple sclerosis". Grounded in this multi-hop chain (ibudilast → CTRP Terminology → multiple sclerosis), the model revises its answer to yes. The example underscores GraphRAG’s advantages: explainability (the supporting path is inspectable), multi-hop reasoning (chained relations guide decoding), and parsimony (a small, high-signal subgraph suffices).
In summary, GraphRAG delivers ~6–7\% relative accuracy lifts over a strong generative baseline without additional training, and yields 0.37\% to 11.78\% relative gains over parametric UMLS injection on the same QA benchmarks. Beyond accuracy, it provides transparent evidence, supports multi-hop reasoning, and enables fast knowledge refresh by updating the graph rather than the model, features that are particularly valuable for biomedical systems whose knowledge base evolves rapidly.
\section{Conclusion and Future Work}
We presented two complementary pathways for injecting structured biomedical knowledge into language models from a single curated source (UMLS-2024AA): parametric injection via continual pretraining on a textualized graph, and non-parametric injection via GraphRAG, which consults the same graph at inference. Using one consistent knowledge base allowed a clean comparison of how knowledge is accessed, embedded in parameters versus retrieved on demand, rather than what knowledge is available.
Empirically, continual pretraining reliably strengthens a general-domain backbone, with the most pronounced benefits on knowledge-intensive tasks, while producing more nuanced, task-dependent effects for an already specialized model. In parallel, GraphRAG improves biomedical QA without updating weights, and brings practical advantages: transparent, inspectable evidence paths; explicit support for multi-hop reasoning through labeled relations; and easy knowledge refresh by updating the graph rather than retraining the model. Together, these findings suggest a pragmatic division of labor: use parametric injection to endow general models with durable domain competence, and deploy inference-time retrieval when explainability, freshness, or operational simplicity are paramount.
This comparison also sets the stage for a broader study of knowledge-injection design choices. In future work, we will systematically benchmark against alternatives such as adapter-based tuning, prompt optimization, and hybrid schemes that align retrieved evidence with parametric updates (e.g., retrieval-aware continual pretraining or distillation of graph context into model weights). We release the processed UMLS Neo4j graph to support reproducibility and to encourage further exploration of structured knowledge for biomedical NLP.

\section{Bibliographical References}\label{sec:reference}
\bibliographystyle{lrec2026-natbib}
\bibliography{lrec2026-example}
\label{lr:ref}
\end{document}